\documentclass{article}

\usepackage[final]{corl_2023} 

\title{Compositional Zero-Shot Learning for Attribute-\\Based Object Reference in Human-Robot Interaction}

\author{
  Peng Gao$^{1*}$, Ahmed Jaafar$^{1*}$, Brian Reily$^2$, Christopher Reardon$^3$, and Hao Zhang$^1$\vspace{3pt}\\ 
 \small $^1$University of Massachusetts Amherst, 
  $^2$DEVCOM Army Research Laboratory,
  $^3$University of Denver\\
  \texttt{\small penggao.robotics@gmail.com, ajaafar@umass.edu, brian.j.reily.civ@army.mil} \\
  \texttt{\small christopher.reardon@du.edu, hao.zhang@umass.edu} \vspace{3pt}\\
  *Authors contributed equally to this paper\\
}
\usepackage{amsthm}

\usepackage{subfigure}
\usepackage{times}
\usepackage{epsfig}
\usepackage{amsmath}
\usepackage{amssymb}
\usepackage{mathrsfs}
\usepackage[ruled,vlined,linesnumbered]{algorithm2e}
\usepackage{mathtools}
\usepackage{lipsum}
\usepackage{graphicx}
\usepackage{wrapfig}
\usepackage{lipsum}

\newcommand{\IMG}{\mathcal{{I}}}
\newcommand{\LLL}{\mathcal{{L}}}

\newcommand{\vv}{\mathbf{v}}

\newcommand{\mm}{\mathbf{m}}

\newcommand{\AAA}{\mathbf{A}}
\newcommand{\COR}{\mathbf{C}}

\begin{document}
\maketitle

\begin{abstract}
    Language-enabled robots have been widely studied over the past years to enable natural human-robot interaction and teaming in various real-world applications. 
    Language-enabled robots must be able to comprehend referring expressions to identify a particular object from visual perception using a set of referring attributes extracted from natural language.
    However, visual observations of an object may not be available when it is referred to, and the number of objects and attributes may also be unbounded in open worlds.
    To address the challenges, we implement an attribute-based compositional zero-shot learning method that uses a list of attributes to perform referring expression comprehension in open worlds. We evaluate the approach on two datasets including the MIT-States and the  Clothing 16K. The preliminary experimental results show that our implemented approach allows a robot to correctly identify the objects referred to by human commands. 
\end{abstract}


\section{Introduction}

\begin{wrapfigure}{R}{0.4\textwidth}
\vspace{-4pt}
\centering
\includegraphics[width=0.38\textwidth]{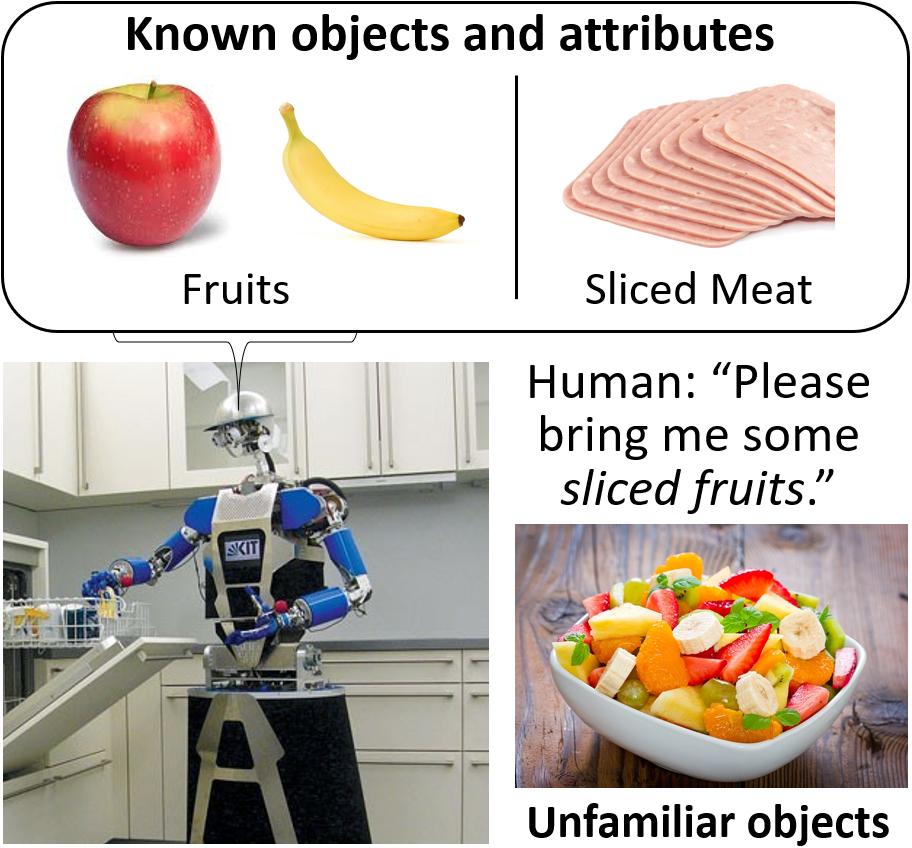}
\caption{A motivating scenario of object reference for human-robot interaction. 
}
\label{fig:motivation}
\vspace{-2pt}
\end{wrapfigure}

Natural language-enabled robots have recently attracted considerable  attention to enable intuitive, efficient, and transparent human-robot interaction and teaming \cite{paul2016efficient,moolchandani2018evaluating,tellex2020robots},
which has a wide variety of real-world applications throughout society, such as in elderly care, hospital assistance, education, inspection, and search and rescue \cite{randall2019survey,bartlett2015human,raman2013sorry}.
Object reference,
defined as the capability of identifying a particular object that a human teammate refers to, is essential for intelligent robots to appropriately communicate with humans \cite{gao2021bayesian}. 
In language-based communication,
a language-enabled robot must comprehend a referring expression in order to recognize and localize a particular object from its visual perception using a set of referring attributes extracted from natural language by the human teammate.



While robot perception has shown promising performance for recognizing object categories, they are insufficient for referring expression comprehension to represent and identify object instances from language and vision. 
First, an object instance can be referred to through language before it is observed, 
and visual data may not be available for the object at the time it is referenced,
meaning that the robot may have to visually identify an unfamiliar object that has not previously observed.
Second, most object recognition and detection techniques focus on identifying the same categories of objects 
but are generally unable to discriminate between object instances, which are more typically referred to. 
Third, in an open world, even given a bounded number of object categories, the number of object instances and the number of object attributes can be unbounded. 
To enable referring expression comprehension and address the above challenges,
we implement an attribute-based compositional zero-shot learning (CZSL) approach 
that composes a list of seen attributes and object labels to identify unseen object-attribute pairs from a robot's perception data in an open world.


\section{Related Work}
\label{sec:citations}
In this study, our primary focus is on compositional zero-shot learning (CZSL) \cite{mancini2021open}. CZSL involves working with a training dataset that comprises various combinations of states and objects. The primary objective in CZSL is to recognize and identify previously unseen combinations of these attributes and objects during testing. We generally divide the existing methods into two groups, including pure visual-based methods and language-prior methods.

For the first group of methods, several approaches have been proposed. Some methods tackle CZSL by separately learning classifiers for objects and states and then combining these classifiers to construct the final recognition model, such as using an SVM classifier trained for known combinations, and class weights for new combinations are inferred using a Bayesian framework \cite{chen2014inferring}. LabelEmbed \cite{misra2017red} introduces a transformation network built on top of pre-trained state and object classifiers. Another work \cite{nagarajan2018attributes} suggests encoding objects as vectors and states as linear operators that transform these vectors. Similarly, a recent work \cite{li2020symmetry} enforces symmetrical representations of objects based on their state transformations. For the second group of methods, the mainstream is to minimize the loss between visual and linguistic learning space for unseen attribute-object pairs \cite{shrivastava2012constrained}. Some recent works use graph structure to leverage information transfer between seen to unseen pairs using graph convolutional networks \cite{mancini2022learning} or graph attention networks \cite{xu2021relation} along with modular networks.

In this paper, 
we implement an attribute-based compositional zero-shot learning (CZSL) approach 
that composes a list of seen attributes and object labels to identify unseen object-attribute pairs from the robot's perception data. The approach allows robots to disentangle attributes and object appearances from observed compositions and predict unseen or unknown compositions in open-world scenarios.
\section{Approach}
\label{sec:citations}

\subsection{Problem Formulation}

We implement an approach for object reference in the context of human-robot interaction. In this scenario, we assume a shared environment containing a human and an assistant robot, such as a living room or a workspace. The primary objective is for the assistant robot to accurately identify the object to which the human is referring, based on the natural language instructions provided.

Specifically, let's consider a typical human referring expression like ``black pen". This expression can be readily parsed by existing natural language models, such as BERT \cite{devlin2018bert}, into its attributes and class label. We denote the attribute ``black" as $w_{attr}$ and the class label ``pen" as $w_{obj}$. 
Based upon the parsed keywords $w_{attr}, w_{obj}$, the robot then undertakes the task of identifying the referred object in the shared environment, drawing upon its own observation sequence, which we denote as $\LLL = \{\IMG_1,\dots,\IMG_n\}$. Formally, we formulate the referring expression comprehension as follows:
\begin{equation}\label{eq:sim}
    s_i = cos\left( \phi(\IMG_i), \psi \left(w_{attr}, w_{obj}\right) \right) = \frac{\phi(\IMG_i)\psi(w_{attr}, w_{obj})}{||\phi(\IMG_i)|| \;||\psi(w_{attr}, w_{obj})||}
\end{equation}
where $\phi$ denotes the network to encode visual features of $\IMG_i \in \LLL$, which is constructed based on a ResNet \cite{he2016deep} or a Vision-Transformer \cite{dosovitskiy2020image} followed by an Average Pooling operation, then the final visual feature is computed through a linear layer.  $\psi$ denotes the network to encode word features of $w_{attr}, w_{obj}$, which is based on GLoVe \cite{pennington2014glove} followed by a multi-layer perception (MLP). $cos$ denotes the cosine function that computes the similarity. $s_i$ denotes the similarity between the query text command and the robot observation. 
Given the similarity, the object reference is formulated as a classification problem, which applies the SoftMax function to the similarity score and predicts which class of attribute-object pairs the observation belongs to, thus achieving object reference.

\subsection{Attribute-Based CZSL}
To make the classification generalizable to unseen attribute-object pairs, based on the existing work \cite{saini2022disentangling, hao2023learning}, we use an attribute-based compositional zero-shot learning method to disentangle attributes of objects for unseen object-attribute pair prediction. Formally, the approach introduces two images $\IMG_{attr}$ and $\IMG_{obj}$ into the learning process. $\IMG_{attr}$ has the same attribute as $\IMG$ but with a different object. $\IMG_{obj}$ has the same object as $\IMG$ but with different attributes. For example, if $\IMG$ is a black pen, then $\IMG_{attr}$ can be a black fork with the same attribute ``black" as $\IMG$ but with a different object class ``fork", denoted as $w_{non-obj}$.  $\IMG_{obj}$ can be a sliver pen with the same object class as $\IMG$ but with a different attribute ``sliver", denoted as $w_{non-attr}$. Our goal is to make a robot not only recognize the black pen that is seen in training but also recognize the silver fork by composing the disentangled attributes and object classes.

Formally, first the correlation between $\IMG, \IMG_{attr}$ and between $\IMG , \IMG_{obj}$ needs to be computed to disentangle the correlated attribute and object class. The correlation matrix is defined as:
\begin{equation}
    \COR^{attr} = \frac{\phi(\IMG_i)\phi(\IMG_{attr})}{||\phi(\IMG_i)||_2 \;||\phi(\IMG_{attr})||_2} 
    \quad and \quad 
    \COR^{obj} = \frac{\phi(\IMG_i)\phi(\IMG_{obj})}{||\phi(\IMG_i)||_2 \;||\phi(\IMG_{obj})||_2} 
    \end{equation}
where $\phi$ denotes the  visual feature encoder based on a ResNet or a Vision Transformer. $\COR^{attr}$ denotes the correlation matrix between $\IMG$ and $\IMG_{attr}$ on the attribute $w_{attr}$, and $\COR^{obj}$ denotes the correlation matrix between $\IMG$ and $\IMG_{obj}$ on the object class $w_{obj}$.
Given the correlation matrix, row-wise and column-wise SoftMax operations are performed on it to get the mask for different attributes and object classes, which is defined as
\begin{equation}
    \AAA_i = \frac{e^{ \COR^{attr}_{i,:}}}{\sum_{i=1}^{d} e^{s_{i,j}}} 
   \quad and \quad 
   \AAA^{attr}_j = \frac{e^{ \COR^{attr}_{:,j}}}{\sum_{i=1}^{d} e^{s_{i,j}}}
   \quad and \quad 
     \AAA^{obj}_j = \frac{e^{ \COR^{obj}_{:,j}}}{\sum_{i=1}^{d} e^{s_{i,j}}}
\end{equation}
where $d$ denotes the dimensions of the feature channels , $ \AAA_i $ and $\AAA^{attr}_j$ denotes the correlation masks between $\IMG$ and $\IMG_{attr}$ in the feature space. $ \AAA^{obj}_j $ denotes the correlated region of $\IMG_{obj}$ on the object class in the feature space. To further disentangle the unseen attributes and object class for generalizable prediction, the following based on the negative correlation matrix is computed:
\begin{equation}
   \AAA^{non-obj}_j = \frac{e^{ -\COR^{attr}_{:,j}}}{\sum_{i=1}^{d} e^{s_{i,j}}}
   \quad and \quad 
   \AAA^{non-attr}_j = \frac{e^{ -\COR^{obj}_{:,j}}}{\sum_{i=1}^{d} e^{s_{i,j}}}
\end{equation}
where $ \AAA^{non-obj}_j $ denotes the non-correlation mask between $\IMG$ and $\IMG_{attr}$ about the object class $w_{non-obj}$.  $\AAA^{non-attr}_j $ denotes the non-correlation mask between $\IMG$ and $\IMG_{obj}$ about the attribute $w_{non-attr}$. To get the final masks, the elements in these masks are added up along their feature channels, which is defined as $m_j = \sum_{i=1}^d \AAA_{ij}$,
where $\mm=\{m_j\}^l$ and $l=49$ denotes the feature length in this paper. Similarly, $\mm^{attr}$, $\mm^{obj}$, $\mm^{non-attr}$ and $\mm^{non-obj}$ can be obtained based on $\AAA^{attr}, \AAA^{obj}, \AAA^{non-attr}, \AAA^{non-obj}$ respectively.
Given the feature masks, the disentangled feature is computed as follows:
\begin{equation}\label{eq:v1}
\vv_{attr} =  \mm \cdot \phi(\IMG_{attr}) + \mm^{attr} \cdot \phi(\IMG),
\end{equation}
\begin{equation}
\vv_{obj} =  \mm \cdot \phi(\IMG_{obj}) + s\mm^{obj} \cdot \phi(\IMG)
\end{equation}
\begin{equation}
\vv_{non-attr} =  \mm^{non-attr} \cdot \phi(\IMG_{attr}),
\end{equation}
\begin{equation}\label{eq:v4}
\vv_{non-obj} =  \mm_{non-obj} \cdot \phi(\IMG_{obj})
\end{equation}
where $\vv_{obj}, \vv_{attr}, \vv_{non-obj}, \vv_{non-attr}$ denote the disentangled features of $w_{obj}, w_{attr}, w_{non-obj}, w_{non-attr}$ respectively.
Given the disentangled features, classification is performed given Eq. (\ref{eq:sim}) by replacing $\phi(\IMG)$ with the disentangled features to predict the object-attribute pair given human text command. Cross entropy loss is used to train the network, in which the loss is minimized between all the visual features defined in Eqs. (\ref{eq:v1}-\ref{eq:v4}) and their associated word embedding features.

\section{Preliminary Experimental Results}
\label{sec:result}

\begin{wraptable}{R}{0.52 \textwidth}
\vspace{-15pt}
\centering
\tabcolsep=0.2cm
\caption{Quantitative Results using the MIT-States Dataset based on the \textbf{AUC} ($\%$) metric.}
\label{tab:quant}
\begin{tabular}{|c|c|c|c|c|}
\hline
  & \textbf{Seen} & \textbf{Unseen} & \textbf{Object} & \textbf{Attr} \\
\hline
Val $@1$ &0.3368 & 0.2965&  0.3683 & 0.3190 \\
\hline	
Val $@2$ &0.4501 & 0.4382&  0.4800 & 0.4531 \\
\hline	
Val $@3$ &0.5206 & 0.5195&  0.5508 & 0.5320 \\
\hline	
Test $@1$ &0.316 & 0.2585&  0.3336 & 0.2844 \\
\hline	
Test $@2$ &0.4252 & 0.3811&  0.4442 & 0.4035 \\
\hline	
Test $@3$ &0.4987 & 0.4577&  0.5160 & 0.4793 \\
\hline	
\end{tabular}
\vspace{-6pt}
\end{wraptable}
We use two existing datasets to evaluate the approach: MIT-States \cite{isola2015discovering} and Clothing16K \cite{zhang2022learning}. 
The MIT-States dataset contains object classes such as fish, rooms, etc. It also has attributes, such as mossy, dirty, etc.
Clothing16K contains clothes, such as suits and pants, with different attributes, such as pink and black. 
In the training stage, all of the objects and attributes are seen. In the testing stage, the compositions of seen and unseen objects/attributes are used to evaluate the generalization of the approach.
In addition, various scenarios are evaluated including seen object-attribute pairs (\textbf{Seen}), unseen object-attribute pairs (\textbf{Unseen}), unseen objects with seen attributes (\textbf{Object}), and seen objects with unseen attributes (\textbf{Attr}).


Area Under the Curve (\textbf{AUC}) is utilized to evaluate the accuracy of text-to-image retrieval.  For each scenario, the top K ($@k$) candidates are retrieved as the results.


Table \ref{tab:quant} presents the quantitative results using the MIT-States dataset. 
From these results, we observe that the approach can successfully retrieve the correct images given the human text commands. The performance of unseen pairs retrieval is slightly worse than the seen pairs, which indicates the generalization capability of the approach for object references in human-robot interaction.
\begin{figure*}[th]
\centering
\subfigure[Evaluation on Clothing16K Dataset]{\includegraphics[height=2.23cm]{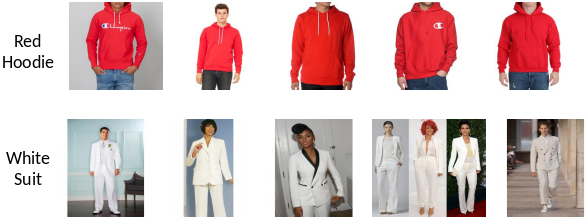}\label{fig:t2i}}
\hspace{6pt}
\subfigure[Evaluation on Real Robot]{\includegraphics[height=2.23cm]{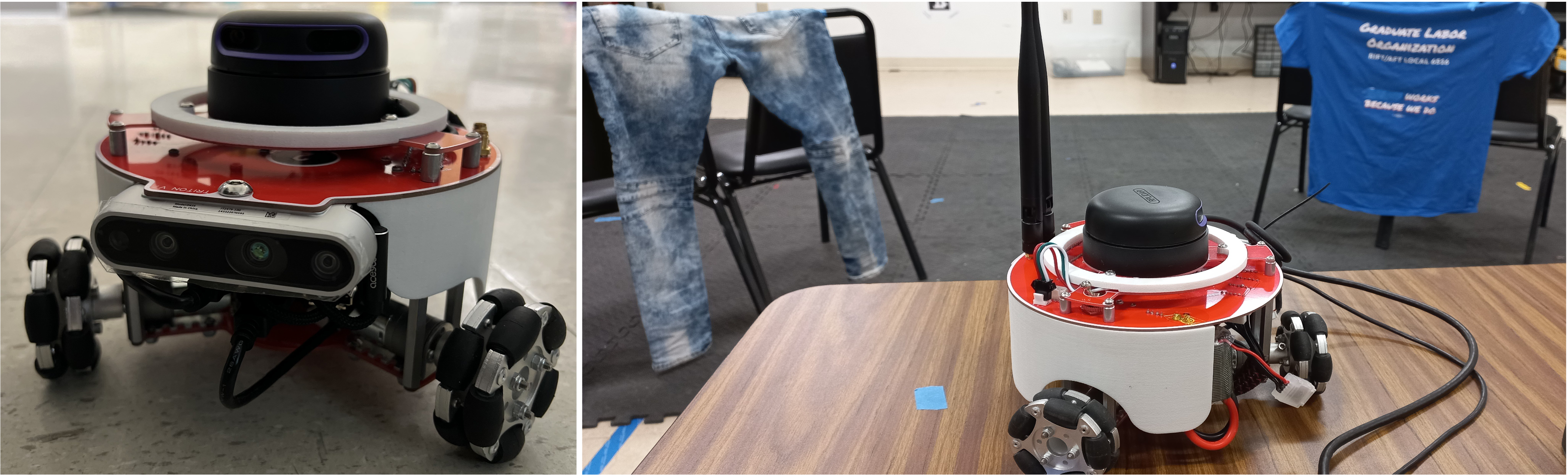}\label{fig:triton}}
\caption{Experiment setups. (a) Retrieval of unseen object-attribute pairs given text query command. (b) Case studies on object reference in human-robot interaction. }
\label{fig:QualResults}
\end{figure*}

We further implement a method based on the recent work \cite{hao2023learning}, it obtains the quantitative results using the Clothing16K dataset, as shown in Figure \ref{fig:t2i}. 
Given the query text command, the approach can correctly retrieve the unseen object-attribute pairs. 
We deploy our implemented work on a physical robot, Triton, to conduct a case study. 
In this study, the Triton moves around its environment to identify the unseen attribute-object pair given via a human textual command. As shown in Figure \ref{fig:triton}, the human command is ``blue shirt" which is unseen in the training stage. 
Eventually, the robot can correctly identify the ``blue shirt" and move forward to it, given that some of the attribute-objects it was trained on are ``red shirt" and ``blue shorts".



\section{Conclusion}
\label{sec:conclusion}
In this paper, 
we implement a new attribute-based compositional zero-shot learning (CZSL) approach to enable referring expression comprehension for human-robot interaction. The approach composes a list of unknown attributes and object labels to identify unseen object-attribute pairs from the robot's perception data in an open world.
Our preliminary experimental results have validated the effectiveness of the approach.

Our current implementation has some limitations. In the future, we will further study the following aspects, including 1) integrating natural language processing with the vision network to perform human-robot interaction tasks, 2) adding robot search and navigation modules to complete the loop, 3) doing a deeper analysis with more sophisticated robots in real-world experiments.



\acknowledgments{This research was partially supported by the ONR grant N00014-21-1-2418, the ARL A2I2 Program W911NF-23-2-0005, and the NSF CAREER Award IIS-2308492.}

\bibliography{camera_ready}  

\end{document}